\documentclass[onecolumn]{article}
\usepackage{graphicx}
\usepackage{float}

\begin{document}

\title{Backpropagation Through Time For Networks With Long-Term Dependencies}

\author{
    Author: \\
    Bird, George$\;\;$ \\
    \texttt{george.bird@student.manchester.ac.uk}
    \and
    Editor: \\
    Polivoda, Maxim E.$\;\;$ \\
    \texttt{mxp19tkp@bangor.ac.uk}
}

\date{10 January 2021}

\maketitle

\begin{abstract}
    \textbf{Note: This paper was written when I was an early undergraduate and I did not realise this was a rediscovery of Real Time Recurrent Learning (RTRL) \cite{RTRL}. Hence, please refer to the original RTRL manuscript instead.}
    \\\\
    Backpropagation through time (BPTT) is a technique of updating tuned parameters within recurrent neural networks (RNNs). Several attempts at creating such an algorithm have been made including: Nth Ordered Approximations and Truncated-BPTT. These methods approximate the backpropagation gradients under the assumption that the RNN only utilises short-term dependencies. This is an acceptable assumption to make for the current state of artificial neural networks. As RNNs become more advanced, a shift towards influence by long-term dependencies is likely. Thus, a new method for backpropagation is required. We propose using the 'discrete forward sensitivity equation' and a variant of it for single and multiple interacting recurrent loops respectively. This solution is exact and also allows the network's parameters to vary between each subsequent step, however it does require the computation of a Jacobian.  
\end{abstract}

\section{Introduction}
Recurrent neural networks (RNNs) are a form an iterative process by feeding forward information from one state of the network into the next time-step. This produces problems when performing backpropagation as all parameters which contributed to the output state must be taken into account, including all those in the previous iterations of the network. Often, it is only information produced recently, from not too many time-steps ago, which helps determine the output of the RNN for the current time-step. For example, this could be the previous few syllables it has classified in an RNN designed for speech recognition. If this is the case the RNN is said to only have short-term dependencies. However, it is likely that as technology advances there will be a shift towards using details across a longer time period. For example, more contextual information may become important for classifying the meaning of speech. In this case the network is said to be using a range of short and long-term dependencies.

Although, if these long-term dependencies influence the current time-step of the RNN then their contribution must be factored into the backpropagation algorithm. Implementing this crudely could be done by saving all previous states to memory and propagating the gradients through every single state before updating the parameters. However, this is not feasible in-the-limit where there are infinite number of iterations in the chain, thus requiring both infinite memory and processing time. Therefore, there have been several attempts at more-efficiently approximating the true gradients.

\subsection{Nth Ordered Approximation}
A simple method, would be to suggest that these recurred gradients are negligible, compared to the contributions by the current state of the network. Thus the \emph{recurred gradients} are simply ignored in their entirety. This is called a $0^{th}$ order approximation. By ignoring these contributions only the recurring cell states and the current iteration of the network need to be stored in memory and backpropagated through to update the tuned parameters - this is considerably less computationally demanding.

Yet the assumption of negligible contributions is not a good assumption to make, since the suggestion is akin to saying that the recurrent feed-back has negligible effect on the output of the network. In effect it is equivalent to reducing a RNN to a fully connected network. Alike performances are not observed between fully connected networks and RNNs, so this assumption is invalid. 

Therefore, one can choose an arbitrary number of previous iterations, which have a plausible contribution to the current network state to perform backpropagation through. This is $N^{th}$ order approximations. Consequently, \emph{N} versions of the network must be loaded into memory at any one time and require approximately \emph{N} times as many calculations to update the tuned parameters as the $0^{th}$ order. In $\lim_{N\to\infty}$ it returns to an exact solution for backpropagation through time. A balance must be struck between N steps and the size of the network, for what can be practically be computed in an acceptable amount of time, putting a limit on the size and/or performance of the RNN.

\subsection{Truncated Backpropagation Through Time}
Truncated Backpropagation Through Time (Truncated-BPTT)\cite{IlyaSutskever}, is a method to segment the recurrency chain into smaller partitions and calculate the respective gradients for each partition. This is a refined version of $N^{th}$ order approximations. In Truncated-BPTT the gradients are only updated after a forward pass through the partition's entire sequence, instead of updating gradients at every iterative step. This has the advantage of being much quicker to evaluate than $N^{th}$ order approximations, at the expense of not updating the network's parameters between each step. We shall call this backpropagating over a constant network. This has the same memory demands as $N^{th}$ order approximations, yet only has the computational demands of approximately the $0^{th}$ order approximation. Hence, is the most widely used technique for BPTT.

Despite its wide usage, there are two key compromises in this method. The first is that the network's parameters are not updated between each time-step - it is kept in a constant parameter state. Consequently, it is an approximation of the desired BPTT algorithm - where the ideal solution would allow the network's parameters to vary between each time-step. Secondly, the partition is limited to a short, constant, finite length. Hence, this technique assumes that the network does not utilise any dependencies from time-steps which fall outside the chain. Thus, the network can only achieve short-term dependencies of fewer time-steps than the size of the partition. This is more detrimental for time-steps earlier in the partition which are assumed to have even shorter dependencies.

\section{Exact Iterative-BPTT Method}
We now propose a new method for calculating BPTT using the discrete forward sensitivity equation\cite{ForwardSensitivityEquation} shown in $Eqn. 1$. This method produces the \emph{exact} gradients as opposed to the previously approximated ones. In addition, it has comparable memory and processing demands to forward-accumulating auto-differentiation on a $1^{st}$ order approximation scenario. This method accounts for all previous time-steps of the RNN, therefore all possible dependencies are factored into the update gradients and no assumptions are needed. The number of network time-steps is continually increasing with each forward pass of the network, yet the computation required for this method of BPTT remains constant - so is ideal for networks expected to run for many time-steps. This would not ordinarily be true for a naive implementation of BPTT, where complexity grows linearly with the number of steps (this is why Truncated-BPTT introduces a 'cut-off' in the partition size). Moreover, the network's parameters are allowed to vary between every time-step, thus can be updated by gradient descent at every time-step.

The drawback is that several 2D-Jacobians must be calculated as well as a Jacobian multiplication - however the latter is well optimised now. The length of each dimension is given by the number of network parameters and the neuronal width of the recurrent loop respectively - this is no different from a forward accumulating method. Since the Jacobian is of constant size, the associated computational complexity to compute this method of backpropagation remains constant irrespective of how many time-steps have contributed to the current RNNs behaviour. This is an initially large computation, however for networks with many time-steps and RNNs expected to be influenced by long term dependencies, then this method is more favourable than the previously mentioned approximate methods - as in these conditions the assumptions for the approximate methods are not valid. To clarify, since this method is an exact representation of the gradients, it handles all dependency time-spans. However, if the RNN is expected to only utilise short-term information it would be better to continue using the approximate methods.

Specifically this new method will have the computational advantage when the number of recurrent time-steps grows large enough that the computation required for naive BPTT is larger than the computation required to compute the Jacobian in $Eqn. 1$. Hence, this method should be used when it is expected that the RNN will be running for many time steps and utilise information from an iteration long ago. Using this method, it would now be feasible to design a network to have a long-term memory based on feed-back loops.

The procedure is achieved by iterating a function of the gradients, whilst a forward pass of the network is computed. This is the discrete forward sensitivity equation, shown below in $Eqn. 1$:

\begin{equation}
    \label{eqn:ForwardSensitivityEquation}
    \Delta_{N} = \frac{\mathrm{d} R_N}{\mathrm{d} P_{N}} + \frac{\mathrm{d} R_N}{\mathrm{d} R_{N-1}} \Delta_{(N-1)}
\end{equation}

Only this Jacobian and the recurred outputs must be stored between successive time-steps. Using $Eqn. 1$, BPTT is \emph{exactly} described by $Eqn. 2$: 

\begin{equation}
    \label{eqn:FSEingradientdescent}
    \frac{\mathrm{d} Y_N}{\mathrm{d} P} = \frac{\mathrm{d} Y_N}{\mathrm{d} P_{N}} + \frac{\mathrm{d} Y_N}{\mathrm{d} R_{(N-1)}}\Delta_{(N-1)}
\end{equation}

Then standard updating by gradient descent is described in $Eqn. 3$:

\begin{equation}
    \label{eqn:FSEinparameterupdate} 
    P_{N+1} = P_{N} - \eta \frac{\mathrm{d} C}{\mathrm{d} P} = P_{N} - \eta \frac{\mathrm{d} C}{\mathrm{d} Y_N} \frac{\mathrm{d} Y_N}{\mathrm{d} P}
\end{equation}

Where $\Delta_{N}$ is the iterated  variable which is stored on the $N^{th}$ time-step. $R_N$ is the vector of recurred values also stored for the next iteration. $P_{N}$ is the set of tunable network parameters on the $N^{th}$ time-step. $C$ is the network cost. Derivation of $Eqn. 1$ is shown in the background section.

Due to current artificial RNNs not exhibiting sufficiently long-term dependencies, the advantages of this method will not be reflected well by results on any currently testable model. In fact, it is expected to perform poorer, on computation time, than the approximate methods for networks only using short-term dependencies due to the approximation's assumptions. However, the constant computation complexity required for successive time-steps can be seen by inspection of this new method and this is a significant improvement over previous attempts which linearly scaled with the number of time-steps. Hopefully, this is sufficiently indicative for good performance on networks for when they transition to utilising more long-term dependencies.

\subsection{Adaptation For Multiple Recurrencies}
Many neural networks are not singularly dependent on one recurrent loop. Models such as an Long Short Term Memory (LSTM) network are dependent on several different recurrent loops. We will index these feed-back loops as $\left\{R^k : \forall k\in\left\{1,2,...,K\right\}\right\}$ and $\bar{K}=\left\{1,2,3,...,K\right\}$ - for K recurrent loops. This will now require a unique $\Delta_{N}$ for each recurrent loop. Thus, we shall now define an adaptation of the forward sensitivity equations required for multiple interacting feed-back loops.

Unfortunately, the formula for multiple recurrent loops is more complex due to the potential interactions between different recurrent loops. Consequently, cross terms must now be accounted for in the 'Delta Function'. Accounting for these cross terms the amended delta function is shown in $Eqn. 4$. The parameter update rules remain the same as described in $Eqn. 3$.

\begin{equation}
    \label{eqn:AdaptedForwardSensitivityEquation2}
    \Delta^k_{N} = \frac{\mathrm{d} R^k_N}{\mathrm{d} P_{N}} + \sum_{l\in\bar{K}}\frac{\mathrm{d} R^k_N}{\mathrm{d} R^l_{N-1}} \Delta^l_{(N-1)} : \forall k \in \bar{K}
\end{equation}

It can be shown that if $\bar{K}=\left\{0\right\}$ then $Eqn. 4$ reduces to $Eqn. 1$. Likewise $Eqn. 2$ requires updating for multiple recurrencies as shown in $Eqn. 5$.

\begin{equation}
    \label{eqn:AdaptedFSEingradientdescent}
    \frac{\mathrm{d} Y_N}{\mathrm{d} P} = \frac{\mathrm{d} Y_N}{\mathrm{d} P_{N}} + \sum_{k \in \bar{K}} \frac{\mathrm{d} Y_N}{\mathrm{d} R^k_{(N-1)}}\Delta^k_{(N-1)}
\end{equation}

The first term in $Eqn. 4$ (and $Eqn. 1$) reperesents standard backpropagation - this is most extensivly used in feed-forward networks. The second term is the required correction for recurrent networks. Therefore, using $Eqn. 4$ and $Eqn. 5$, one can derive the standard backpropagation algorithm and all the approximate algorithms for backpropagation through time. So the combination of $Eqn. 4$ and $Eqn. 5$ could be considered a more general algorithm - where the backpropagation algorithm and approximate BPTT algorithms are for specific circumstances.

The ways in which artificial neurons connect can be seperated into three fundamental categories: feed-forward, feed-back and terminating. A feed-forward connection is one which is in the direction of input neurons to output neurons and is contained within the \emph{directed acyclic graph}. A feed-back loop is described by a collection of connections which can cause an artificial neuron to influence itself at a later time period - each feed-back loop must contain at least one feed-back connection. Therefore, feed-back connections are those which when eliminated from the network's directed graph leave a directed acyclic graph in the direction from input neurons to output neurons. Feed-forward and feed-back connections must be able to influence the output neurons too. Finally a terminating connection is any connection which cannot influence any output neuron in any way. 

Each of these connection types has an associated backpropagation algorithm. Standard backpropagation can be considered the associated solution for any purely feed-forward artificial neural network whilst BPTT is the associated solution for a more general artificial neural network which can also include feed-back connections. Since terminating connections do not influence the output neurons then they do not require a backpropagation algorithm. Since these are the three fundamental directional connections and backpropagation is described exactly by the first term in $Eqn. 4$ whilst BPTT is described exactly by $Eqn. 4$, then we propose that $Eqn. 4$ and $Eqn. 5$ may together describe a full solution of backpropagation for any given artificial neural network.

On inspection of $Eqn. 4$ and $Eqn. 5$ it can be seen that the gradient complexity follows $O\left(k^2\right)$ due to the number of cross terms to be calculated for each interacting recurrent loop. However, this can be further simplified by analysis of some fundamental properties of a recurrent neural network.

\subsubsection{Recurrency-Abstraction Heuristic}
Mathematically, we define abstraction as decreasing the intrinsic dimensionality of a data manifold. A data manifold is described in the "Manifold Hypothesis" \cite{ChrisOlah}. Abstraction qualitatively manifests as collating many elements into one object class. In neural networks this is done by clustering points in a manifold towards one another. A perfect abstraction would be the collapse of part of a manifold into an intrinsically lower dimension - regardless of what the extrinsic dimensionality of the system is. This, process is irreversible. Now, we can define the extrinsic dimensionality of the system by the number of neurons in a specified layer of the system - where activations of neurons are represented in the space and each individual neuron serves as a basis vector. These distributed activations form the manifolds. So we can mandate an abstraction to occur between layers, by forcing the extrinsic dimensionality of the next layer to be lower than the intrinsic dimensionality of the manifold. Thus, we incur irreversible collapses of the manifold in several locations - in effect information is lost in the transformation.

A bidirectional neuron can be imagined. It is able to propagate activations in either direction (up and down subsequent layers) and the layers are fully connected in this scenario. If it propagates to a layer which has a higher extrinsic dimension, the manifold is simply embedded at some angle in the space with its intrinsic dimensionality preserved. Whilst, if the layer has a extrinsic dimensionality less than the intrinsic dimensionality then abstraction occurs. Thus, the transformations are naturally asymmetric - resulting in an effective directionality principle for any neural network where a manifold's intrinsic dimensionality varies. Unlike these bidirectional networks, artificial neural networks are inherently directional anyway. 

But for any recurrence, we can show that the process is abstractive for a given manifold. A recurrence occurs when a layer '$L$' of a network is fed-back and concatenated with a previous layer '$L-k$'. Thus, the extrinsic dimensionality of the layer '$L-k+1$' (given by $Ex(L-k+1)$) where '$L-k+1$' is the layer given by the concatenation of $L$ and $L-K$, must be greater than the extrinsic dimensionality of the fed-back layer $L$ written as: $Ex\left(L-k+1\right)>Ex\left(L\right)$. The concatenation, joins the two manifolds thus summing their intrinsic dimensionality, written as: $In\left(L\right)+In\left(L-k\right) = In\left(L-k+1\right)$. We know that the intrinsic dimensionality on the later fed-back layer is just one component of this sum and since intrinsic dimensionality cannot be negative then the intrinsic dimensionality of layer '$L$' must be less than or equal to the intrinsic dimensionality of layer '$L-k+1$' written as: $In\left(L-k+1\right) \geq  In\left(L\right)$. Consequently, given that neither the intrinsic dimensionality of the manifold in layer '$L$' or in layer '$L-k$' is zero, the process must be abstractive. Therefore, the mapping is surjective but non-injective, resulting in a directional, non-invertible, process. This conclusion is reperesented by $Eqn. 6$. Note this is strictly for recurrent neural networks which use a concatenation when feeding back information. 

\begin{equation}
    \label{eqn:IntrinsicDimensionalityRelation}
        In\left(L-k+1\right) \neq 0 \cap In\left(L\right) \neq 0 \Rightarrow In\left(L-k+1\right) > In\left(L\right)
\end{equation}

Due to this, the interaction between recurrent loops must be directional too. Thus, if the gradients in one interaction direction are non-zero, then the inverted interaction's gradients must be zero. Therefore, cancelling $50\%$ of the cross terms - they needn't ever be calculated.

\subsection{Temporal Gradient Explosion}
Phenomena observed in typical feed-forward networks are expected to have temporal analogoues in feed-back networks. Just as gradient explosions can plague deep neural networks, we can now expect to observe \emph{temporal gradient explosions} - the equivalent phenomena for recurrent neural networks. This occurs if the gradient sum is divergent. This could be accounted for by adding an attenuation factor for each iterative gradient to force a convergence, however this will once again make the method an approximation not an exact solution. On the opposite side, we now ideally would want the effect of \emph{temporal gradient vanishing} as this would converge the sequence.

Gradient explosion is characterised in the forward pass as small perturbations becoming increasingly larger, so we can predict that the network becomes chaotic as time progresses. It remains deterministic but small changes in a value quickly grow out of control - potentially impeding the function of the network. This principle is the \emph{'butterfly effect'} and perhaps recurrent neural networks with this new backpropagation technique may have some success in being able to model chaotic dynamical systems.

\section{Background}
The following is a derivation of the discrete forward sensitivity equation for artificial neural networks.

We wish to calculate the full time-dependent gradients for every parameter for a recurrent neural network in its $N^{th}$ iterative state. A basic recurrent neural network is outlined in $Fig. 1$.

\begin{figure}[H]
    \begin{center}
        \includegraphics[width=0.9\textwidth]{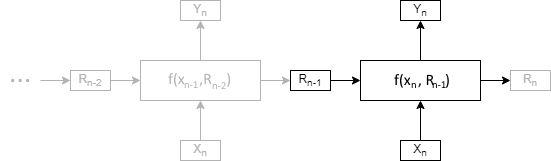}
        \caption{Recurrent Neural Network, on time-step N, produces output Y as a function of inputs X and feed-back input $R_{n-1}$ and tuned parameters $P_{n}$ individually indexed by $\forall i\in I$ as shown by $P_n = \{ P_{i,n} : \forall i \in I\}$.}
    \end{center}
\end{figure}

The full gradients for this network, with respect to the output layer, are described by $Eqn. 7$ and summarised in $Eqn. 8$. Gradient descent with respect to error $C$ is shown in $Eqn. 9$ with learning rate $\eta$. It can be seen that as N becomes large the backpropagation becomes impractical to calculate since there are N terms to consider. This is the linear scaling of computational complexity for successive time-steps using a naive BPTT approach.

\begin{equation}
    \label{eqn:RecurrentBackprop} 
    \frac{\mathrm{d} Y_N}{\mathrm{d} P} = \frac{\mathrm{d} Y_N}{\mathrm{d} P_{N}} + \frac{\mathrm{d} Y_N}{\mathrm{d} P_{N-1}} + \frac{\mathrm{d} Y_N}{\mathrm{d} P_{N-2}} + ... + \frac{\mathrm{d} Y_N}{\mathrm{d} P_{0}}
\end{equation}
\\
\begin{equation}
    \label{eqn:RecurrentBackpropSummarised} 
    \frac{\mathrm{d} Y_N}{\mathrm{d} P}= \sum_{n=0}^{N}\frac{\mathrm{d} Y_N}{\mathrm{d} P_{n}}
\end{equation}
\\
\begin{equation}
    \label{eqn:RecurrentGradientsDescent} 
    P_{N+1} = P_{N} - \eta \frac{\mathrm{d} C}{\mathrm{d} P} = P_{N} - \eta \frac{\mathrm{d} C}{\mathrm{d} Y_N} \frac{\mathrm{d} Y_N}{\mathrm{d} P}
\end{equation}
\\
Next begin with expanding $Eqn. 7$, using chain rule, into  $Eqn. 10$, then summarising $Eqn. 10$ into $Eqn. 11$.

\begin{equation}
    \label{eqn:RecurrentBackpropIterativlyExpressed}
    \frac{\mathrm{d} Y_N}{\mathrm{d} P} =
\frac{\mathrm{d} Y_N}{\mathrm{d} P_{N}}+
\frac{\mathrm{d} Y_N}{\mathrm{d} R_{N-1}}\frac{\mathrm{d} R_{N-1}}{\mathrm{d} P_{N-1}} +
\frac{\mathrm{d} Y_N}{\mathrm{d} R_{N-1}}\frac{\mathrm{d} R_{N-1}}{\mathrm{d} R_{N-2}}\frac{\mathrm{d} R_{N-2}}{\mathrm{d} P_{N-2}} + ... +
\frac{\mathrm{d} Y_N}{\mathrm{d} R_{N-1}}\left(\prod_{n=0}^{N-2}\frac{\mathrm{d} R_{n+1}}{\mathrm{d} R_{n}}\right)\frac{\mathrm{d} R_{0}}{\mathrm{d} P_{0}}
\end{equation}
\\
\begin{equation}
    \label{eqn:RecurrentBackpropIterativlyExpressedSummarised}
    \frac{\mathrm{d} Y_N}{\mathrm{d} P} = \frac{\mathrm{d} Y_N}{\mathrm{d} P_{N}} + \frac{\mathrm{d} Y_N}{\mathrm{d} R_{N-1}}\frac{\mathrm{d} R_{N-1}}{\mathrm{d} P_{N-1}} + \sum_{n=0}^{N-2}\frac{\mathrm{d} Y_N}{\mathrm{d} R_{N-1}}\left( \prod_{k=n}^{N-2}\frac{\mathrm{d} R_{k+1}}{\mathrm{d} R_{k}}\right)\frac{\mathrm{d} R_{n}}{\mathrm{d} P_{n}}
\end{equation}
The first term in $Eqn. 11$ is the existing backpropagation for feed-forward neural networks. The second and third terms are the required corrections for any neural networks with a feed-back loop. 

We can now reformulate $Eqn. 11$ as a recurrence relation, to perform at each time-step with the forward propagation of the network as shown in $Eqn. 12$. We need only now store and calculate the variable $\Delta_{N}$ at each time-step. This final reformulation is the discrete forward sensitivity equation:

\begin{equation}
    \label{eqn:FSE}
    \Delta_{N} = \frac{\mathrm{d} R_N}{\mathrm{d} P_{N}} + \frac{\mathrm{d} R_N}{\mathrm{d} R_{N-1}} \Delta_{(N-1)}
\end{equation}

For multiple interacting feed-back loops, the discrete forward sensitivity equation shown in $Eqn. 12$ must be adapted to $Eqn. 13$. Each feed-back loop is indexed by superscript $k$. $Eqn. 13$ is the adapted discrete forward sensitivity equation:

\begin{equation}
    \label{eqn:AdaptedForwardSensitivityEquation1}
    \Delta^k_{N} = \frac{\mathrm{d} R^k_N}{\mathrm{d} P_{N}} + \sum_{l\in\bar{K}}\frac{\mathrm{d} R^k_N}{\mathrm{d} R^l_{N-1}} \Delta^l_{(N-1)}
\end{equation}

Therefore, backpropagation through time is \emph{exactly} described by $Eqn. 14$. 

\begin{equation}
    \label{eqn:Iterative}
    \frac{\mathrm{d} Y_N}{\mathrm{d} P} = \frac{\mathrm{d} Y_N}{\mathrm{d} P_{N}} + \frac{\mathrm{d} Y_N}{\mathrm{d} R_{(N-1)}}\Delta_{(N-1)}
\end{equation}


\begin{thebibliography}{9}
    \bibitem{RTRL} Williams, Ronald J., and David Zipser. ``A learning algorithm for continually running fully recurrent neural networks.'' Neural computation 1.2 (1989): 270-280.

    \bibitem{IlyaSutskever} 
    Ilya Sutskever, ``Training Recurrent Neural Networks". \textit{Graduate Department of Computer Science, University of Toronto}, 2013.

    \bibitem{ForwardSensitivityEquation} 
    Christopher Rackauckas, Yingbo Ma, Vaibhav Dixit, Xingjian Guo, Mike Innes, Jarrett Revels, Joakim
    Nyberg, and Vijay Ivaturi. "A Comparison of Automatic Differentiation and Continuous
    Sensitivity Analysis for Derivatives of Differential Equation Solutions".  arXiv preprint arXiv:1812.01892, Dec 2018.

    \bibitem{ChrisOlah}
    Chris Olah, "Neural Networks, Manifolds, and Topology".\\ \textit{https://colah.github.io/posts/2014-03-NN-Manifolds-Topology/} Accessed 10/01/2021.
\end{thebibliography}
\end{document}